\documentclass[letterpaper, 10 pt, conference]{ieeeconf}  

\IEEEoverridecommandlockouts                              

\usepackage{graphics} 
\usepackage{graphicx}
\usepackage{url}
\usepackage{amsmath}
\usepackage{amsfonts}
\usepackage[table]{xcolor}
\usepackage{listings}
\usepackage{subcaption}
\usepackage{multirow}
\usepackage{siunitx}
\usepackage{placeins}
\usepackage{algorithm}
\usepackage{algorithmicx}
\usepackage{algpseudocode}
\usepackage{amssymb}
\usepackage{bm}
\usepackage{bbm}
\usepackage{color}
\usepackage{hyperref}
\newcommand{\be}{\begin{equation}}
\newcommand{\ee}{\end{equation}}
\newcommand{\bd}{\begin{displaymath}}
\newcommand{\ed}{\end{displaymath}}
\newcommand{\BE}{\begin{eqnarray}}
\newcommand{\EE}{\end{eqnarray}}

\definecolor{arsenic}{rgb}{0.23, 0.27, 0.29}
\definecolor{charcoal}{rgb}{0.21, 0.27, 0.31}
\definecolor{hanblue}{rgb}{0.27, 0.42, 0.81}
\definecolor{blue-ncs}{rgb}{0.0, 0.53, 0.74}
\definecolor{awesome}{rgb}{1.0, 0.13,0.32}
\definecolor{darkgreen}{rgb}{0, .4,0}

\title{\LARGE \bf
	A Study on the Use of Simulation in Synthesizing Path-Following Control Policies for Autonomous Ground Robots
}

\author{Harry Zhang$^{1}$, Stefan Caldararu$^{1}$, Aaron Young$^{2}$, Alexis Ruiz$^{1}$, Huzaifa Unjhawala$^{1}$,Ishaan Mahajan$^{1}$, \\Sriram Ashokkumar$^{1}$,  Nevindu Batagoda$^{1}$, Zhenhao Zhou$^{1}$, Luning Bakke$^{1}$, and Dan Negrut$^{1}$
\thanks{$^{1}$University of Wisconsin-Madison, Madison, WI 53706, USA
        {\tt\small \{hzhang699,scaldararu,aruiz26,unjhawala, imahajan,ashokkumar2, batagoda,lfang9,zzhou292, negrut\}@wisc.edu}}%
\thanks{$^{2}$Massachusetts Institute of Technology, Cambridge, MA 02139, USA
        {\tt\small aryoung@mit.edu}}%
\thanks{$^{3}$Autonomy Research Testbed Open-Source Platform: https://github.com/uwsbel/autonomy-research-testbed}
}

\begin{document}

\maketitle
\thispagestyle{empty}
\pagestyle{empty}


\begin{abstract}
We report results obtained and insights gained while answering the following question: how effective is it to use a simulator to establish path following control policies for an autonomous ground robot? While the quality of the simulator conditions the answer to this question, we found that for the simulation platform used herein, producing four control policies for path planning was straightforward once a digital twin of the controlled robot was available. The control policies established in simulation and subsequently demonstrated in the real world are PID control, MPC, and two neural network (NN) based controllers. Training the two NN controllers via imitation learning was accomplished expeditiously using seven simple maneuvers: follow three circles clockwise, follow the same circles counter-clockwise, and drive straight. A test randomization process that employs random micro-simulations is used to rank the ``goodness'' of the four control policies. The policy ranking noted in simulation correlates well with the ranking observed when the control policies were tested in the real world. The simulation platform$^{3}$ used is publicly available and BSD3-released as open source; a public Docker image is available for reproducibility studies. 

\end{abstract}

\section{INTRODUCTION}
\label{sec:intro}
Simulation is touted as having the potential to accelerate the synthesis of control policies and the design of perception and planning algorithms~\cite{PNASsimRobotics2021}. Designing in simulation can reduce costs, accelerate the design cycle, and improve safety during the design and after the solution is deployed in the real world. There are several examples of combining simulation with recent advances in Reinforcement Learning (RL), Imitation Learning (IL), and model based control that have yielded encouraging results in autonomous driving, see, for instance,~\cite{bewley2019learning,Osinski2020SimbasedRL,kalapos2020RLpathfollowing,hamilton2022ZST}. These and similar policy synthesis efforts can rely on existing simulators, e.g., \cite{carlaAVsim2017,gazebo,webots2004,airsim2018,chronoOverview2016,isaacNVIDIA}. However, as documented herein, there are remarkably few instances in which a contribution reports on a simulation-based design while also providing a full account of the policy's behavior upon deployment on an actual robot; i.e., reporting on the sim2real gap \cite{sim2realGapEssex1995}. In many cases, there is no real-world testing, and the policy synthesized in simulation is validated in the same simulator. This is undesirable because latent biases in a simulator shape the autonomy solution, and testing in the same simulator will not expose the solution's brittleness. Furthermore, the community benefits little from efforts that do not close the loop and provide real-world feedback, which is crucial for improving the quality of the simulators.

This contribution reports results obtained and lessons learned while answering the following question: how useful is it to employ a simulator to synthesize path-following control policies for a scale ground robot? To this end, we use a digital twin of a 1/6th scale vehicle called ART (from Autonomous Research Testbed vehicle) \cite{artatk2022}, and synthesize in a simulator four control policies: a PID controller, an MPC solution, and two NN controllers. These control policies are tested in a series of 60 experiments carried out using ART on a grassy area and on top of a parking lot to assess the sim2real gap. 

We acknowledge that there are many simulators and a horizontal study that compares their effectiveness for the task at hand would be insightful. This is not the goal of this contribution, which instead uses an in-house developed simulator \cite{chronoOverview2016} and control policy design support therein, to provide empirical evidence that a simulator can provide, zero-shot style, control policies for path following. Moreover, the scope of the study had to be narrowed again in relation to the task considered in this exercise. Path following is a fundamental task in robotics, and the interest here is to gauge the ability of a policy synthesized in simulation to work in reality as well, when drawing on IMU and RTK-GPS sensing.  We are interested in whether one can accomplish zero-shot transfer, that is, getting similar performance in reality by using \textit{with no modification} the policy synthesized in simulation. Finally, the insights gained are translational -- using a superior simulator to the one employed would only improve the conclusions reached in this study.

\section{RELATED WORK}
\label{sec:relatedWorkd}
The problem of interest in this study is that of controlling a vehicle to follow a path defined by waypoints. A survey of the literature suggests that efforts in this area fall in one of three categories: (a) reality-only policy design, in which the design effort takes place only in the real world; (b) simulation-only policy design that trains and validates the policies only in simulation without closing the loop with real-world demonstration; and (c) simulator-to-reality policy design, when a policy is first synthesized and tested in simulation and then evaluated in real world. 

\medskip

\noindent{\textbf{Reality-Only Policy Design}}: An event-triggered Model Predictive Control (MPC) for solving the path following problem has been shown to perform well in indoor environments when drawing on data from an Inertial Measurement Unit (IMU) and an indoor position system~\cite{Rother2023EventTriggeredControl}. In~\cite{pan2020imitation}, the authors designed a path following policy using Imitation Learning, where the ``expert'' was in fact an MPC policy. The solution relied on GPS, IMU, wheel decoders, and camera sensors. More recently, an online learning coupled with an MPC for path following using GPS/IMU sensors has been demonstrated to be effective~\cite{wang2023GPlearningandMPC}. Therein, integrating online learning with an MPC policy improves the path following performance of the solution. From a high vantage point, the ``reality-only'' design approach is costly and time intensive to collect training data or tuning policy in reality environments.  

\medskip

\noindent{\textbf{Simulation-Only Policy Design}}: In~\cite{jin2023pathfollowing}, the authors propose an H-infinity state feedback control policy for path following tasks that outperforms the traditional Linear Quadratic Regulator (LQR) in MATLAB and CarSim~\cite{benekohal1988carsim} simulation. In~\cite{sierra2024RLpathfollowing}, the authors designed a path following policy using RL and demonstrated the effectiveness of the policy in a simulated environment by comparing it to a PID controller. In~\cite{zhang2021adaptiveMPCpathfollowing}, the authors proposed an adaptive MPC policy and demonstrated its effectiveness over a traditional MPC algorithm using a simulation environment without resorting to tests in the real world. On the upside, the ``simulation-only'' approaches are expeditious since one can easily test a control policy. On the downside, these solutions are brittle once deployed in reality, although domain randomization \cite{domainRandomizationAbbeel2017} can help in this regard.

\medskip

\noindent{\textbf{Simulator-to-Reality Policy Design}}: Most of the work in the field of path following sim2real transfer uses ML or RL related algorithms. In~\cite{kalapos2020RLpathfollowing,hamilton2022ZST}, an RL policy based on input from a monocular camera sensor (or a stereo camera with a LiDAR) is trained in simulation first and then deployed in the real world. In~\cite{voogd2023reinforcementpathfollowing}, the authors successfully transferred their RL policy based off GPS and IMU sensors from simulation to reality. Lastly, in~\cite{kovacs2023optimizationtubecontrol}, the authors present both simulation and reality experimental results for model predictive tube control, but simulation is used more as a ``validation'' step as opposed to a control policy-design facilitator. 

\medskip

Herein, we are interested in the ``Simulator-to-Reality Policy Design'' category, and our contributions are twofold. First, we add to the body of knowledge by documenting results of a study that focuses on the sim2real transferability issue in robotics. The study is specific to one simulator, Chrono~\cite{chronoOverview2016}, and one task, path following. We show that it is exceedingly easy to synthesize four stock policies in simulation that mitigate the sim2real gap, zero-shot style. Second, we contribute an approach that proved effective in ranking the real-world behavior of control policies synthesized in simulation. It is anchored by \textit{test randomization}, and promotes the use of random micro-simulations that provide a statistical perspective on the robustness and effectiveness of a controller. The control policy ranking noted at the conclusion of the test randomization process in simulation carried over to the ranking of the control policies when deployed in reality. In this context, the idea of test randomization is complementary to the idea of domain randomization \cite{domainRandomizationAbbeel2017}. While test randomization helps with ranking competing control policies, domain randomization helps with synthesizing and hardening a policy. The work also aligns with the idea outlined in \cite{kadian2020sim2real}, where the authors proposed the sim2real correlation coefficient that correlates behavior in sim and real. However, the methodology proposed here is a-priori, i.e., it is exercised before deploying in the real world. This is unlike the approaches in \cite{kadian2020sim2real,aaronAmesSafetySim2Real2023}, which resort to using real world data to characterize the sim2real gap.

\section{METHODOLOGY}
\label{sec:method}
We use a simulator to verify, tune, and train four stock control policies. The policy ``verified'' in simulation is an MPC solution. The policy ``tuned'' in simulation is a PID controller. Finally, we train two NN solutions through IL. Subsequently, these four control policies (called herein MPC, PID, NN-MPC, and NN-HD) are tested in reality, zero-shot style, to understand how they compare against each other, and whether their performance in reality comes close to their efficacy in simulation; i.e., assess the sim2real gap. The interest is in understanding the role that simulation plays in synthesizing them, the effort to accomplish this, and the level of success these control policies enjoy when deployed in the real world. Note that there is no claim on novelty in relation to the four stock control policies synthesized in this work. As such, the description of the MPC and PID control policies is provided in the Appendix, with the discussion focusing on the NN-type controllers.

\begin{figure}
    \centering
    \includegraphics[width=0.49\textwidth]{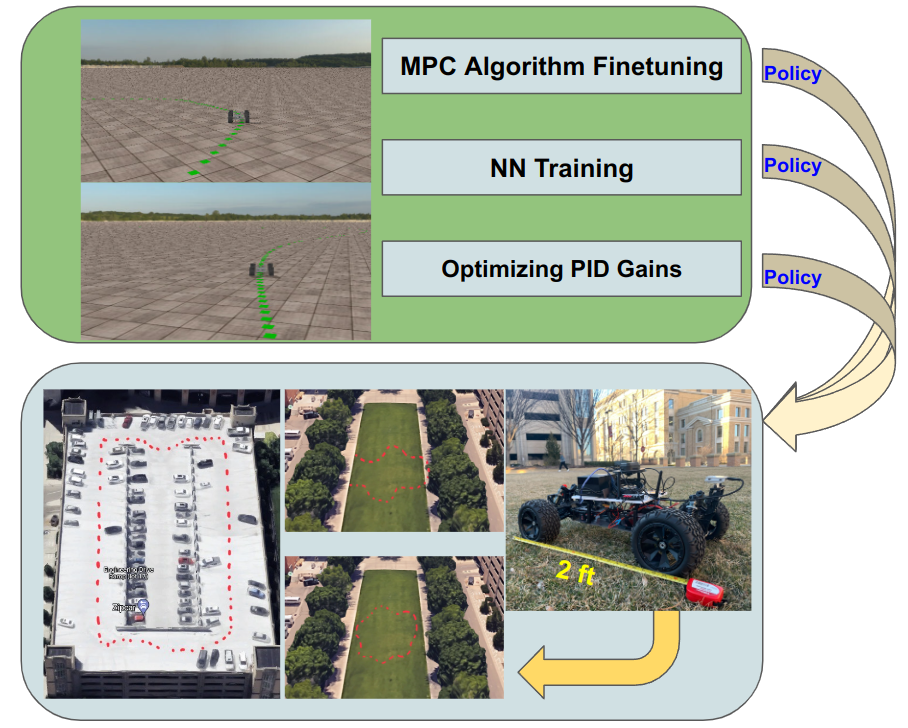}
    \caption{Study overview: MPC, NN and PID policies verified, trained, or tuned in simulation (upper box); policies subsequently tested in reality using a 1/6th vehicle (lower box).}
    \label{fig:pipeline}
\end{figure}

The vehicle platform used in this exercise is described in \cite{artatk2022,TR-2023-15ArtOak}. The simulator is discussed briefly at the end of this section. One GPS (SparkFun GPS-RTK2) and one IMU (WHEELTEC N100) are used. The pipeline that anchors this methodology is illustrated in Fig.~\ref{fig:pipeline}.

\subsection{Imitation Learning-Based Neural Network Controller (NN-HD, NN-MPC)}
As mentioned in the Appendix, the error state $\mathbf e \in \mathbb R^{4 \times 1}$ for the vehicle is a low dimensional input for the problem, which makes it suitable for accurate and cost-effective NN training. The following two subsections introduce the data collection and training process for NN-HD and NN-MPC. The reason for using two different training experts is to investigate the sim2real transferability, generalizability, and the sim2real gap itself (since these two policies were designed with the same training pipeline). For the Human Driver (HD) expert, the benefit lies in its minimal expertise requirement, free from vehicle modeling, control problem setup, or any pre-designed policy. Conversely, in~\cite{pan2020imitation}, it was shown that the MPC expert is attractive for its accurate optimal behavior.

\begin{figure}[ht]
	\centering
	{{\includegraphics[width=0.49\textwidth]{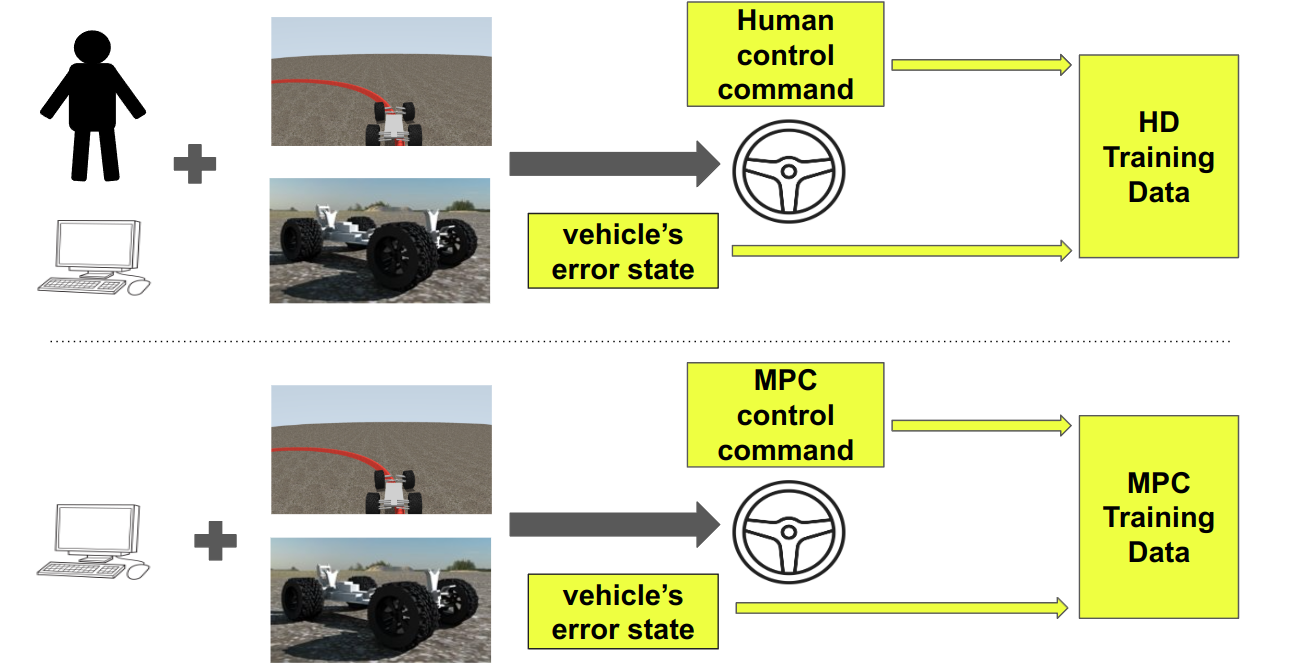} }}%
	\caption{Handling of the NN-based policies: upper half illustrates the approach for collecting NN-HD training data, which is done with a human in the loop; lower half shows data collection process for NN-MPC, which relies on an MPC expert, following the idea in \cite{pan2020imitation}.}
	\label{fig:training}
\end{figure}

\noindent{\textbf{Human Driver produced training data}}. This approach is schematically captured in Fig.~\ref{fig:training}. Since the simulator supports human-in-the-loop (HIL) simulation, a human drives in the simulated virtual world while data is collected that registers what the driver does when the digital twin vehicle strays away from a given trajectory. We record the error state and the corresponding control commands at each time step of the simulation. Seven reference trajectories were used for the training process. The reference trajectories consists of circular paths with radius 2\unit{m}, 5\unit{m}, 25\unit{m}, both clockwise and counter-clockwise, plus a 30\unit{m} straight line path, which can be thought of as a circle with infinite radius. Collecting training data in simulation was both simple and fast -- it took on the order of minutes to generate the training data.

\noindent{\textbf{MPC produced training data}}. Instead of having a human driving in the simulator, we used the MPC controller and recorded the input commands issued by the MPC while it worked to maintain a predefined trajectory. This idea of using an ``MPC expert'' is discussed in \cite{pan2020imitation}. The MPC produces commands to the digital twin vehicle to make it follow the same reference trajectories used in the HD data collection. The training data contains the error state and corresponding MPC command issued in each time step.

We employed and trained a feed forward Neural Network (NN) that will be tested on the digital twin vehicle and subsequently the real vehicle in a zero-shot sim2real transfer. Due to the relatively small dimension of the input and output data, it is sufficient to use a two-hidden-layer NN for this supervised learning problem (hidden layer one: 8 neurons; hidden layer two: 16 neurons). Training data is generated via HD only, or via MPC only. For NN training, the input data $\mathbf{E} \in \mathbb{R}^{4 \times n}$ is the set of error states $\mathbf{e} \in \mathbb{R}^{4 \times 1}$. This data is matched with the corresponding control command $\mathbf{u} \in \mathbb{R}^{2 \times 1}$ from the set $\mathbf{U} \in \mathbb{R}^{2 \times n}$. The NN is trained to produce a mapping between the error state and control command, $\mathbf{f} : \mathbf{e} \rightarrow \mathbf{u}$. The NN training and inference is carried out using Keras Core~\cite{chollet2015keras}, using PyTorch as a backend~\cite{paszke2017PyTorch}. The training converges fast since the input and output spaces are small, as are the NN's depth and width. The average inference time for NN based controller was around $1.8 \unit{ms}$.

\subsection{Simulation Platform}
The simulation platform used is anchored by the open-source, BSD3-released simulator Chrono \cite{projectChronoGithub,pyChronoCondaWebSite,benatti2022pychrono}. Chrono provides support for ground vehicle on- and off-road mobility studies. A terramechanics simulation can take place at several levels of fidelity, and comes with ancillary sensing of terrain deformation. Chrono has templetized vehicle models \cite{chronoVehicle2019}, and it works in conjunction with a ROS2 autonomy stack \cite{artatkResearchPlatform2022}, which allows one to use the same stack, running on the same processor, both in simulation and on the real autonomous vehicle. The sensor models includes GPS, IMU, LiDAR, encoder and camera. LiDAR and camera sensor simulation utilize ray tracing techniques to provide high-fidelity physically-based rendering~\cite{asherSensorSimulation2021}. The infrastructure supports human-in-the-loop simulation (used here to generate training data for NN-HD), and multi-agent scenarios \cite{synchrono2020}. For reproducibility studies, all simulations discussed here can be run using the Docker image associated with this contribution~\cite{HarryIROSPathFollowing}.

\section{EXPERIMENTS \& ANALYSIS}
\label{sec:exp}
We carry out testing both in simulation and the real world. The vehicle uses the GPS to find its location; the IMU is only used for heading. In simulation, the GPS sensor model has access to privileged location information. The same can be said about the heading from IMU, which is provided by virtue of having access to the system's states in the simulator. Although for two trajectories in the real world the vehicle moved over a grassy area, at the time of the testing, the soil was frozen and a rigid representation of the terrain was deemed adequate. The simulation ran in soft real time. The digital twin had previously been calibrated to fit the physical counterpart~\cite{huzaifaIEEE-AccessCalibration2024}. In real world testing, the RTK-GPS position (with $\pm$ 2 \unit{cm}) and IMU heading data was provided to the control policy. Finally, the same longitudinal controller was used both in simulation and real world to maintain a 1~\unit{m/s} speed.

\subsection{A Test-Randomization Approach to Ranking Control Policies in Simulation}
\label{sec:simRanking}
\begin{figure}
    \centering
    \includegraphics[width=0.485\textwidth]{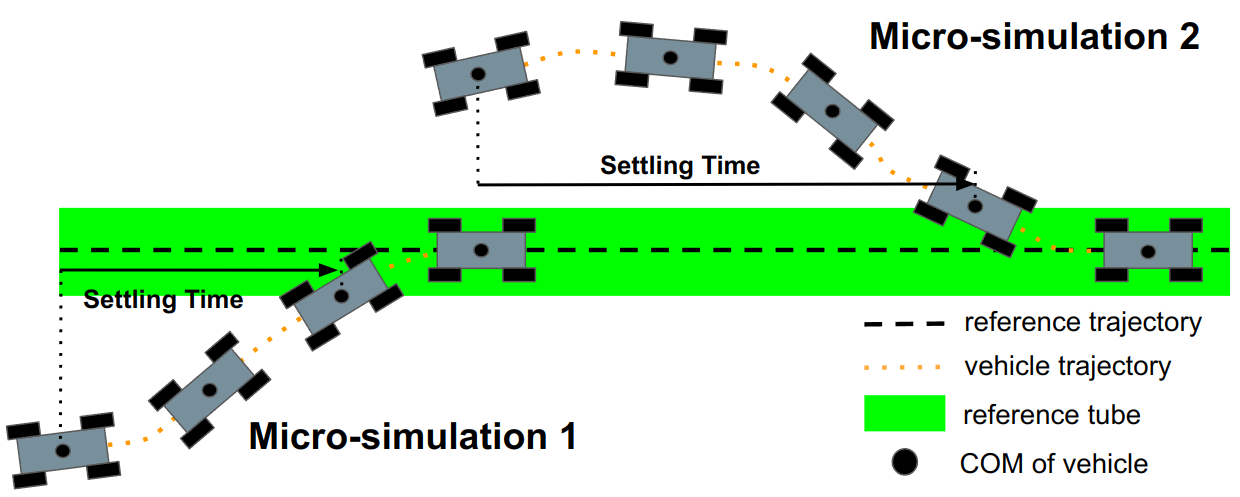}
    \caption{Micro-simulations: green area stands for a schematic of the ``reference tube'' (note that vehicle orientation also defines a tube, not shown here); the figure shows two micro-simulations with different settling time lengths.}
    \label{fig:micro_sim}
\end{figure}
We compare the zero-shot performance of four stock control policies synthesized in the simulator -- MPC, NN-MPC, PID, and NN-HD, see Section \ref{sec:method}. To that end, we run micro-simulations that speak to how easy it is for the vehicle to come back to a straight trajectory when its location and orientation are slightly perturbed by $e$ and $\phi$, i.e., its location is $e$ away from the reference straight trajectory, and it is mis-headed by an angle $\phi$. The $e$ and $\phi$ are realizations of two random variables that are uniformly distributed, $|E| \sim \mbox{U}[1.5,2.5]$ and $\Phi \sim \mbox{U}[-\frac{\pi}{4},\frac{\pi}{4}]$, respectively. In such a micro-sim, if $e=-1.5$ and $\phi = \frac{\pi}{6}$, the digital twin is below the straight line by 1.5 meters, and it is pointed at an angle of $\frac{\pi}{6}$ away from the straight horizontal line (positive angles measured counterclockwise). Once the digital vehicle is placed at an off-location and misdirected, it starts from rest. We measure how long it takes the vehicle to merge back into the ``reference tube'' defined by $|e|,|\phi| < 0.1$. This is called ``settling time'' (ST) (see Fig.~\ref{fig:micro_sim} for more intuitive demonstration of micro-simulations). Thus, a micro-simulation typically runs for a short 10 to 15 seconds. During the micro-simulation, the longitudinal controller works toward getting the longitudinal speed to be 1.0 \unit{m/s}. We draw 100 random pairs of realizations $e_1^p$ through $e_{100}^p$, and $\phi_1^p$ through $\phi_{100}^p$, for each policy $p \in \{\mbox{MPC}, \mbox{NN-MPC}, \mbox{PID}, \mbox{NN-HD}\}$. 

The resulting 400 ST values are used to produce the results in Table~\ref{tab:torture_tests}, which reports how many times each control had the shortest ST, the second shortest ST, etc. The last column indicates the average ST for each policy when run in simulation. Thus, on average, it took MPC 5.471 seconds to settle into the reference tube. For NN-HD, the least competitive policy, it took 9.37 seconds to settle. The results in Table~\ref{tab:torture_tests} suggest a clear winner -- MPC, and a clear laggard -- NN-HD. The performance of NN-MPC is better than that of PID, yet they came close (NN-MPC is better at about 60\% of the time). Finally, the 400 simulations run to reach the conclusion that MPC $>$ NN-MPC $\gtrsim$ PID $>$ NN-HD in terms of performance. It took approximately 2.5 hours to complete 400 micro-simulations. 

\begin{table}[h]
    \centering
    \rowcolors{3}{}{gray!15}
    \resizebox{0.485\textwidth}{!}{
    \begin{tabular}{|l|cccc|c|}
    \hline
    \multirow{2}{*}{Policy} & \multicolumn{4}{c|}{\rule{0pt}{2.5ex}ST Rank} & \multirow{2}{*}{Mean ST} \\
    & 1st & 2nd & 3rd & 4th & \\ 
    \hline 
    \rule{0pt}{2.5ex}MPC     & 98  & 0   & 0   & 2   & 5.471 s \\
    NN-MPC  & 2   & 57  & 36  & 5   & 6.575 s \\
    PID     & 0   & 42  & 58  & 0  & 6.591 s \\
    NN-HD   & 0   & 1   & 6   & 93  & 9.37 s \\ \hline
    \end{tabular}
    }
    \caption{Test Randomization results, suggesting that MPC $>$ NN-MPC $\gtrsim$ PID $>$ NN-HD.}
    \label{tab:torture_tests}
\end{table}

\subsection{Real World Policy Ranking: Sim2Real Transferability}
\label{sec:sim2real}

\begin{figure}[h!]
	\centering
	\includegraphics[width=0.49\textwidth]{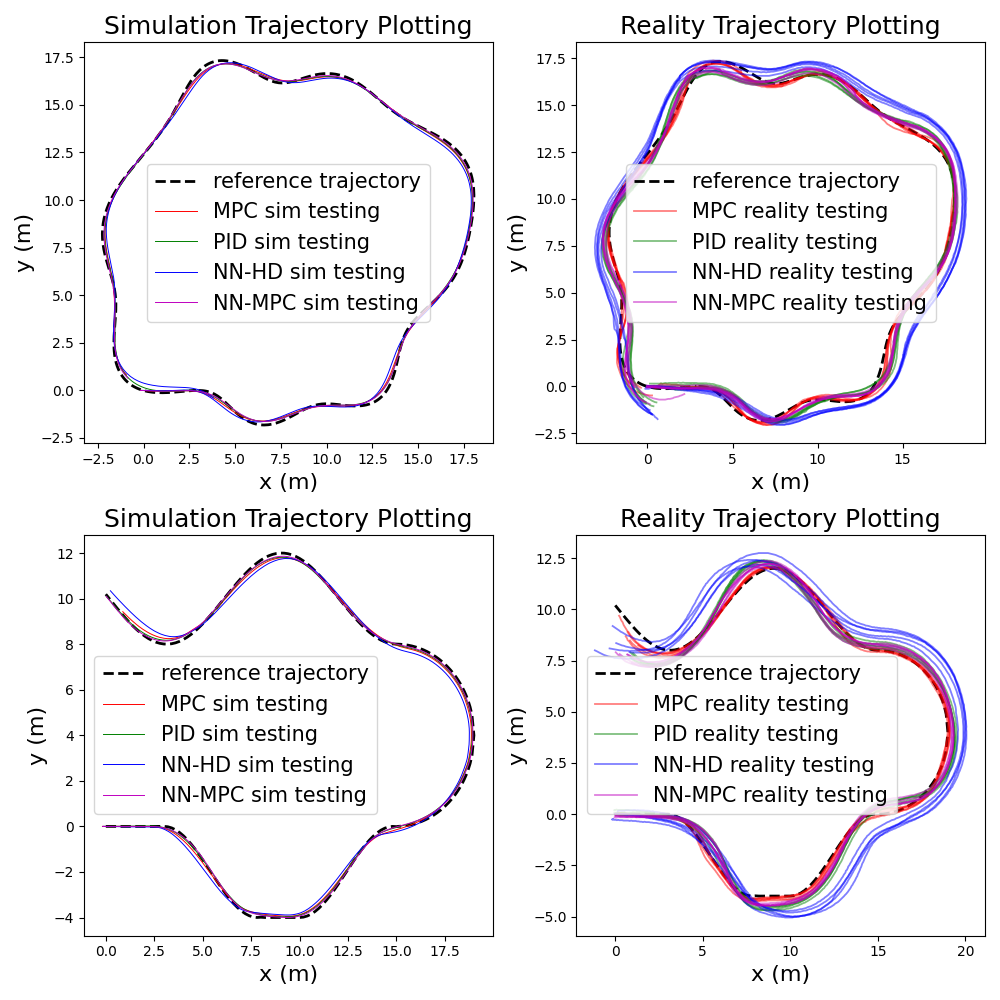}
	\caption{Left column: sample \textit{simulation} trajectories associated with Path 1 (top) and Path 2 (bottom). Right column: sample \textit{real-world} trajectories associated with Path 1 (top) and Path 2 (bottom). See Tables~\ref{tab:lateral_error} and \ref{tab:heading_error} for quantitative information regarding the performance of the MPC, NN-MPC, PID, and NN-HD control policies in sim and real.}
	\label{fig:ECEgrass}
\end{figure}

\begin{figure}[h!]
	\centering
	\includegraphics[width=0.49\textwidth]{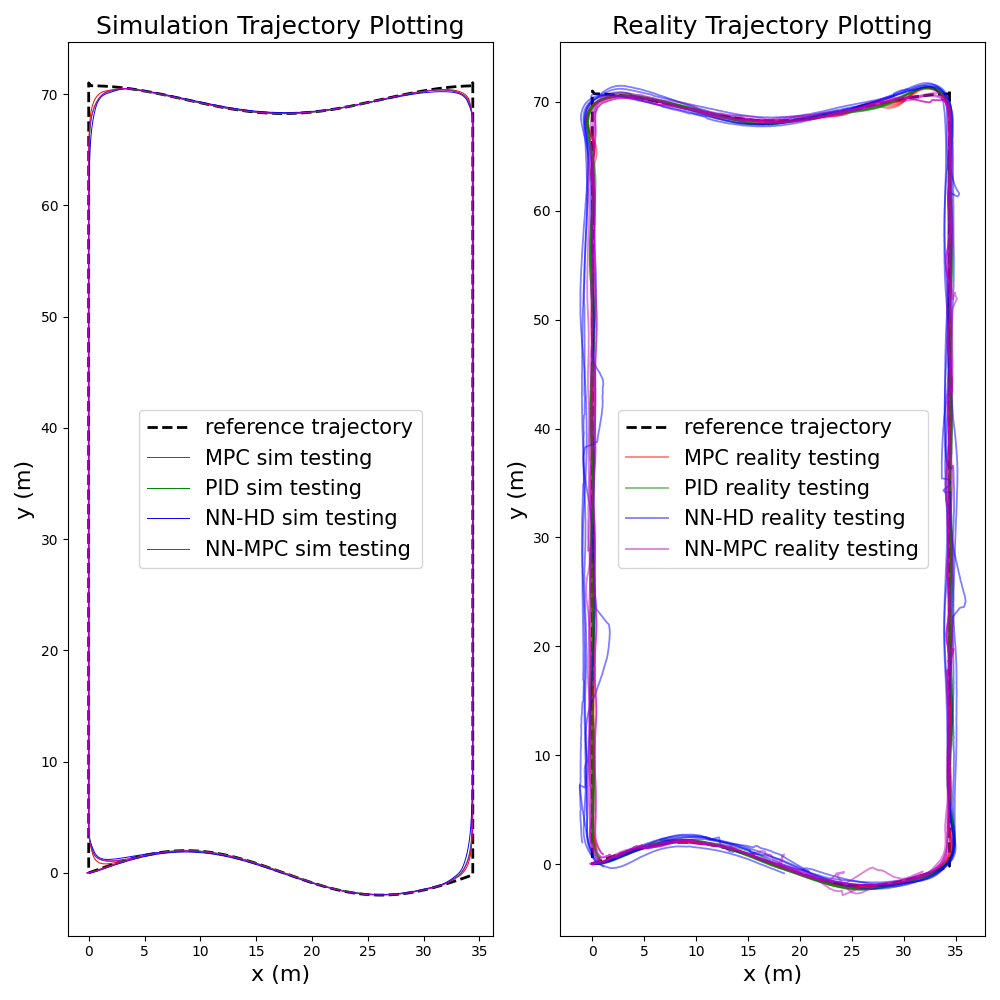}
	\caption{Left column: sample \textit{simulation} trajectories associated with Path 3. Right column: sample \textit{real-world} trajectories associated with Path 3. See Tables~\ref{tab:lateral_error} and \ref{tab:heading_error} for quantitative information regarding the performance of the MPC, NN-MPC, PID, and NN-HD control policies in sim and real.}
	\label{fig:ParkingLot17}
\end{figure}

A second experiment was carried out to answer the following questions. Question Q1: does the MPC $>$ NN-MPC $\gtrsim$ PID $>$ NN-HD ranking noted in simulation translate to the real world? Question Q2: how big is the gap between the behavior of the vehicle in reality, and that noted for its digital twin in simulation? In other words, Q1 asks if the \textit{test randomization} is a good predictor of \textit{relative} performance, and Q2 asks whether there is a gap between sim and real. 

We answer Q1 and Q2 by running 60 experiments in the real world, and running corresponding simulations. We used three paths defined by waypoints. Path 1: a randomly distorted circular path of length 63.5 meters -- see first row in Fig.~\ref{fig:ECEgrass}. Path 2: a 49.5 meter long path made up of a double lane change trajectory, followed by a half circle, followed by a sinusoidal portion -- see second row in Fig.~\ref{fig:ECEgrass}. Path 3: a 212.9 meters long trajectory -- the 70 meter long edges were straight lines, the shorter edges were one a sinusoidal, the other one a sector of a circle, see Fig.~\ref{fig:ParkingLot17}. Paths 1 and 2 were on a grassy area; Path 3 was on concrete, at the top of a parking lot (see also Fig.~\ref{fig:pipeline} for an aerial view of the grassy area and the top of the parking lot). Swapping control policies to test them was exceedingly simple for two reasons. 
First, since our autonomy stack is ROS2 based, swapping a control policy amounted to changing a ROS2 node.
Second, the same autonomy stack \cite{atk-art2022}, with the exact same code, was used both in sim and real. 
To gain a statistical perspective on the real world performance, the scale vehicle used the MPC policy to run each path five times. The same was done for the vehicle using NN-MPC, PID, and NN-HD. This added up to 60 experiments in the real world. The same Path 1, 2, and 3 experiments were subsequently run in simulation, but only once, owing to the deterministic nature of the simulator. 

\begin{figure}[h!]
    \centering
    \includegraphics[width=0.49\textwidth]{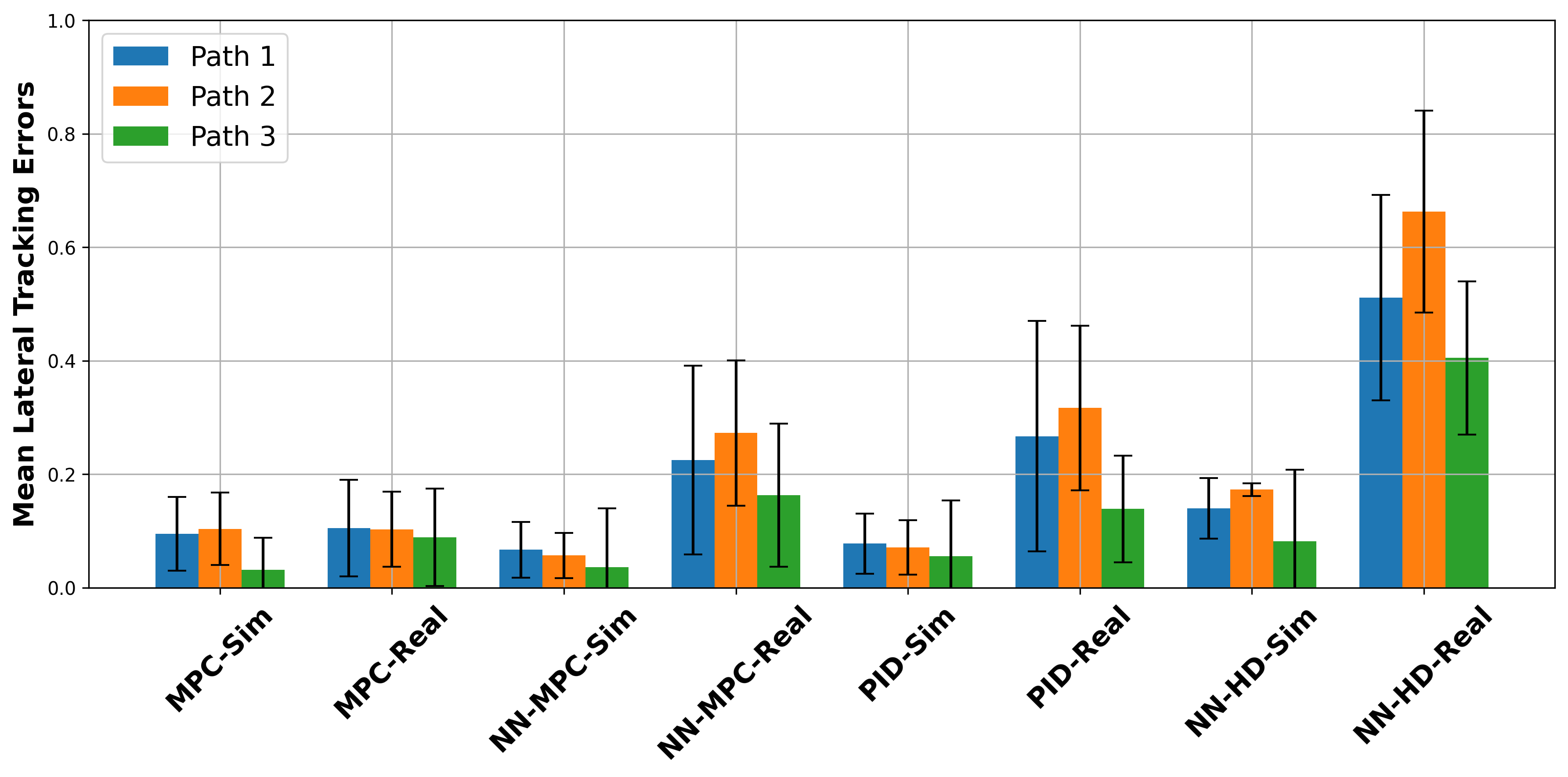}
    \caption{Lateral tracking error comparison between different policies, trajectories, and testing environments.}
    \label{fig:lateral_error}
\end{figure}

In both simulation and the real world, all four policies were able to see through the path following task for all trajectories. However, it is insightful to consider the average lateral tracking and heading errors -- see values in Tables~\ref{tab:lateral_error}~\&~\ref{tab:heading_error}. An error-bars plot for the results in Table~\ref{tab:lateral_error}, displaying average and standard deviation information, is shown in Fig.~\ref{fig:lateral_error}.

\begin{table}[h!]
    \centering
    \resizebox{0.49\textwidth}{!}{
    \rowcolors{3}{}{gray!15}
    \begin{tabular}{|l|c|c|c|}
    \hline
    \multirow{2}{*}{Policies} & \multicolumn{3}{c|}{Lateral Tracking Error in Meters $(\mu \pm \sigma)$}  \\ \cline{2-4}
                              & path 1 & path 2 & path 3 \\ \hline
    MPC-Sim    & $0.095\pm0.065$ & $0.104\pm0.064$ &$0.032\pm0.086$\\ \hline
    MPC-Real   & $0.105\pm0.085$ & $0.103\pm0.066$ &$0.089\pm0.086$\\ \hline
    NN-MPC-Sim & $0.067\pm0.049$ & $0.057\pm0.040$ &$0.036\pm0.104$\\ \hline
    NN-MPC-Real& $0.225\pm0.166$ & $0.273\pm0.128$ &$0.163\pm0.126$\\ \hline
    PID-Sim    & $0.078\pm0.053$ & $0.071\pm0.048$ &$0.056\pm0.098$\\ \hline
    PID-Real   & $0.267\pm0.203$ & $0.317\pm0.145$ &$0.139\pm0.094$\\ \hline
    NN-HD-Sim  & $0.140\pm0.053$ & $0.173\pm0.011$ &$0.082\pm0.126$\\ \hline
	NN-HD-Real & $0.511\pm0.181$ & $0.663\pm0.178$ &$0.405\pm0.135$\\ \hline
    \end{tabular}
    }
    \caption{Lateral tracking error comparison between different policies in simulation and real-world conditions: for each time step, the lateral tracking error is calculated as the shortest distance from the vehicle to the reference trajectory. For one data point in the table, the mean and standard deviation of the lateral tracking error for all 5 repeated tests are presented.}
    \label{tab:lateral_error}
\end{table}

\begin{table}[h!]
    \centering
    \resizebox{0.49\textwidth}{!}{
    \rowcolors{3}{}{gray!15}
    \begin{tabular}{|l|c|c|c|}
    \hline
    \multirow{2}{*}{Policies} & \multicolumn{3}{c|}{Heading Error in Radius $(\mu \pm \sigma)$}  \\ \cline{2-4}
                              & path 1 & path 2 & path 3 \\ \hline
    MPC-Sim    & $0.067\pm0.051$ & $0.064\pm0.059$ &$0.034\pm0.111$\\ \hline
    MPC-Real   & $0.094\pm0.069$ & $0.099\pm0.063$ &$0.091\pm0.132$\\ \hline
    NN-MPC-Sim & $0.058\pm0.049$ & $0.051\pm0.057$ &$0.032\pm0.096$\\ \hline
    NN-MPC-Real& $0.158\pm0.115$ & $0.223\pm0.105$ &$0.128\pm0.106$\\ \hline
    PID-Sim    & $0.061\pm0.047$ & $0.056\pm0.057$ &$0.029\pm0.095$\\ \hline
    PID-Real   & $0.184\pm0.140$ & $0.210\pm0.108$ &$0.111\pm0.132$\\ \hline
    NN-HD-Sim  & $0.061\pm0.047$ & $0.075\pm0.064$ &$0.033\pm0.091$\\ \hline
	NN-HD-Real & $0.313\pm0.134$ & $0.663\pm0.178$ &$0.405\pm0.135$\\ \hline
    \end{tabular}
    }
    \caption{Heading error comparison between different policies in simulation and real testing: for each time step, the heading error is calculated as the difference between vehicle's heading angle and reference heading angle. For one data point in the table, the mean and standard deviation of the heading error for all 5 repeated tests are presented.}
    \label{tab:heading_error}
\end{table}

Q1 and Q2 can now be answered based on the results in Tables~\ref{tab:lateral_error} and \ref{tab:heading_error}, along with the error bars in Fig.~\ref{fig:lateral_error}. Thus, for Q1, the real world results for Paths 1 and 2 coincide with the test-randomization ranking: MPC $>$ NN-MPC $>$ PID $>$ NN-HD. For Path 3, we have MPC $>$ PID $>$ NN-MPC $>$ NN-HD. As the Test Randomization results in Table~\ref{tab:torture_tests} suggest, PID and NN-MPC were close therein, and they came out close in the real world testing too. For the heading error, for Path 1, MPC $>$ NN-MPC $>$ PID $>$ NN-HD, while for Paths 2 and 3, MPC $>$ PID $>$ NN-MPC $>$ NN-HD. Again, the winner and the laggard are clear, with NN-MPC and PID contending for the second place. As for Q2, all control policies that worked in simulation, worked zero-shot style in reality. A very large gap is noted for NN-HD.

\section{DISCUSSION}
\label{sec:disc}
This study set out to analyze how effective the use of simulation is in the design of control policies for path-following of ground robots. In this context, a meaningful question to ask is how important are the dynamics engine, the sensor simulator, the autonomy stack, and the digital twin(s). There is some leeway in relation to sensor simulation and the autonomy stack. For instance, in this study we could have used the Chrono GPS and IMU sensor models; instead, we chose to use privileged state information reading the location and orientation of the vehicle straight out of the simulator without using any noise model. For the autonomy stack, we used the same stack both in the simulator and on the real vehicle. One could bypass it in simulation and instead produce code that embeds a control subroutine in the simulator. We have elected to not do that. Since we used a ROS2 autonomy stack, changing the control policy amounted to swapping a ROS2 node. This convenience came at the cost of setting up a ROS2 autonomy stack \cite{artatk2022}, which is challenging. Finally, the process of producing a calibrated digital twin is arduous~\cite{huzaifaIEEE-AccessCalibration2024}. However, once these four components are in place and validated -- the dynamics engine, sensor simulation, autonomy stack, and digital twin -- using them is highly effective. For a knowledgeable individual, generating the ROS2 nodes to implement the stock control policies was a matter of days, and subsequently running the test randomization process to a-priori rank the contenders was a matter of hours. 

The simulator was used in two stages -- first synthesizing and then ranking the controllers. In the first stage, the simulator was used to verify (for MPC), tune (for PID), and train (for NN-MPC and NN-HD) the policy. For ranking, test randomization provided quickly a statistical perspective on the robustness of the contenders. Note that if we ran simulations only on Path 1, Path 2, and Path 3, which are otherwise non-trivial paths, we would have had only a limited perspective on the comparative efficacy of the control policies (based on results in Table.~\ref{tab:lateral_error} \&~\ref{tab:heading_error}). This is because testing in simulation on three paths cannot provide a thorough perspective on a policy's performance -- the control policies are in a ``comfort zone'' (as the reference tube in Fig.~\ref{fig:micro_sim}) and do not experience major perturbations. The test-randomization produced ranking had better correlation with the performance in real world and rendered a component of ``explainability'' to the fact that the ranking correlation was not perfect. The usefulness of test randomization is posited to be a consequence of the broad coverage of scenarios that it brings to bear. Intuitively, no matter what trajectory a vehicle has to follow, the control policy will always be challenged by small perturbations that budge it off the desired path. Not tying the test to a specific path and having many random micro-simulations that conclude quickly are two strengths of the test randomization approach. In this study, while test randomization is tailored to the context of ground vehicle path-following algorithms, the concept of evaluating a policy's quality through simulation experiments that introduce adversity and perturbation has broader applicability across various domains. Furthermore, given that test randomization can accurately forecast real-world performance rankings, it offers the potential to fully conduct policy optimization, parameter adjustments, and other enhancements within a simulated environment.

Although the goal of this contribution was not to design a new control policy or extol the qualities of an existing one, it is nonetheless interesting to compare the goodness of our MPC, NN-MPC, PID, and NN-HD control policies against the performance of other solutions. For instance, an event-triggered MPC path following algorithm is proposed in~\cite{Rother2023EventTriggeredControl}. Tested in an indoor environment, it had a tracking error in the range of $[0.067,0.145]$, which is in the range of MPC. A Gaussian Process based online learning coupled with MPC is proposed for path following in~\cite{wang2023GPlearningandMPC}. Therein, the authors report lateral tracking errors in the range of $[0.05,0.10]$, which is comparable to the stock MPC algorithm implemented in this contribution. An RL path following algorithm~\cite{voogd2023reinforcementpathfollowing} achieved tracking error in the range of $[0.21,0.62]$, which is in the range of NN-MPC and PID.

This contribution stops short of addressing several related questions. It remains to investigate what happens when the scale vehicle is not bound to move with a constant speed of 1.0\unit{m/s}, but can implement an optimal control policy to reduce time to destination. What would be the right test randomization process doing justice to a scenario like this? As a work in progress, this group is in the process of measuring the sim2real gap and gauging the effectiveness of simulation in more complex tasks, e.g., path following combined with object avoidance. Finally, the test randomization process is an a-priori tool. It ranks relative goodness of various control policies, which might themselves be assembled in the simulator, \textit{prior} to deploying them in the real world. However, the method does not quantify the sim2real gap and presently we are not aware of techniques that do so without collecting real world data, as done in \cite{kadian2020sim2real,aaronAmesSafetySim2Real2023}. 

\section{CONCLUSION}
\label{sec:conc}
This empirical study assessed the feasibility of exclusively using simulation to synthesize, zero-shot style, path-following control policies for scale autonomous robots. The prerequisites for this process are a validated simulator and a calibrated digital twin. Using a ROS2 autonomy stack and sensor simulation, while desirable, is not absolutely necessary. The study demonstrates that designing path-following control policies in simulation is expeditious and effective. Finally, we proposed a test randomization methodology, which by running an ensemble of random micro-simulations provided a statistical perspective on the relative performance of the controllers. We found good correlation between the performance ranking produced in the simulator and the one noted in real world testing.

\section*{ACKNOWLEDGMENT}
This work was in part made possible with funding from NSF project CMMI215385.

\clearpage



\bibliographystyle{IEEEtran}
\bibliography{BibFiles/refsSensors,BibFiles/refsMachineLearning,BibFiles/refsAutonomousVehicles,BibFiles/refsChronoSpecific,BibFiles/refsSBELspecific,BibFiles/refsMBS,BibFiles/refsCompSci,BibFiles/refsTerramech,BibFiles/refsFSI,BibFiles/refsRobotics,BibFiles/refsDEM}

\def\cprime{$'$}
\begin{thebibliography}{10}
\providecommand{\url}[1]{#1}
\csname url@rmstyle\endcsname
\providecommand{\newblock}{\relax}
\providecommand{\bibinfo}[2]{#2}
\providecommand\BIBentrySTDinterwordspacing{\spaceskip=0pt\relax}
\providecommand\BIBentryALTinterwordstretchfactor{4}
\providecommand\BIBentryALTinterwordspacing{\spaceskip=\fontdimen2\font plus
\BIBentryALTinterwordstretchfactor\fontdimen3\font minus \fontdimen4\font\relax}
\providecommand\BIBforeignlanguage[2]{{%
\expandafter\ifx\csname l@#1\endcsname\relax
\typeout{** WARNING: IEEEtran.bst: No hyphenation pattern has been}%
\typeout{** loaded for the language `#1'. Using the pattern for}%
\typeout{** the default language instead.}%
\else
\language=\csname l@#1\endcsname
\fi
#2}}

\bibitem{PNASsimRobotics2021}
\BIBentryALTinterwordspacing
H.~Choi, C.~Crump, C.~Duriez, A.~Elmquist, G.~Hager, D.~Han, F.~Hearl, J.~Hodgins, A.~Jain, F.~Leve, C.~Li, F.~Meier, D.~Negrut, L.~Righetti, A.~Rodriguez, J.~Tan, and J.~Trinkle, ``On the use of simulation in robotics: Opportunities, challenges, and suggestions for moving forward,'' \emph{{Proceedings of the National Academy of Sciences}}, vol. 118, no.~1, 2021. [Online]. Available: \url{https://www.pnas.org/content/118/1/e1907856118}
\BIBentrySTDinterwordspacing

\bibitem{bewley2019learning}
A.~Bewley, J.~Rigley, Y.~Liu, J.~Hawke, R.~Shen, V.-D. Lam, and A.~Kendall, ``Learning to drive from simulation without real world labels,'' in \emph{2019 International Conference on Robotics and Automation (ICRA)}.\hskip 1em plus 0.5em minus 0.4em\relax IEEE, 2019, pp. 4818--4824.

\bibitem{Osinski2020SimbasedRL}
B.~Osi{\'n}ski, A.~Jakubowski, P.~Ziecina, P.~Mi{\l}o{\'s}, C.~Galias, S.~Homoceanu, and H.~Michalewski, ``Simulation-based reinforcement learning for real-world autonomous driving,'' in \emph{2020 IEEE international conference on robotics and automation (ICRA)}.\hskip 1em plus 0.5em minus 0.4em\relax IEEE, 2020, pp. 6411--6418.

\bibitem{kalapos2020RLpathfollowing}
A.~Kalapos, C.~G{\'o}r, R.~Moni, and I.~Harmati, ``Sim-to-real reinforcement learning applied to end-to-end vehicle control,'' in \emph{2020 23rd International Symposium on Measurement and Control in Robotics (ISMCR)}.\hskip 1em plus 0.5em minus 0.4em\relax IEEE, 2020, pp. 1--6.

\bibitem{hamilton2022ZST}
N.~Hamilton, P.~Musau, D.~M. Lopez, and T.~T. Johnson, ``Zero-shot policy transfer in autonomous racing: Reinforcement learning vs imitation learning,'' in \emph{2022 IEEE International Conference on Assured Autonomy (ICAA)}.\hskip 1em plus 0.5em minus 0.4em\relax IEEE, 2022, pp. 11--20.

\bibitem{carlaAVsim2017}
A.~Dosovitskiy, G.~Ros, F.~Codevilla, A.~Lopez, and V.~Koltun, ``{CARLA}: {An} open urban driving simulator,'' in \emph{Proceedings of the 1st Annual Conference on Robot Learning}, 2017, pp. 1--16.

\bibitem{gazebo}
Open-Source-Robotics-Foundation, ``A {3D} multi-robot simulator with dynamics,'' \url{http://gazebosim.org/}, accessed: 2022-03-01.

\bibitem{webots2004}
O.~Michel, ``Cyberbotics ltd. webots: professional mobile robot simulation,'' \emph{International Journal of Advanced Robotic Systems}, vol.~1, no.~1, p.~5, 2004.

\bibitem{airsim2018}
S.~Shah, D.~Dey, C.~Lovett, and A.~Kapoor, ``{AirSim}: High-fidelity visual and physical simulation for autonomous vehicles,'' in \emph{Field and service robotics}.\hskip 1em plus 0.5em minus 0.4em\relax Springer, 2018, pp. 621--635.

\bibitem{chronoOverview2016}
A.~Tasora, R.~Serban, H.~Mazhar, A.~Pazouki, D.~Melanz, J.~Fleischmann, M.~Taylor, H.~Sugiyama, and D.~Negrut, ``{Chrono}: An open source multi-physics dynamics engine,'' in \emph{High Performance Computing in Science and Engineering -- Lecture Notes in Computer Science}, T.~Kozubek, Ed.\hskip 1em plus 0.5em minus 0.4em\relax Springer International Publishing, 2016, pp. 19--49.

\bibitem{isaacNVIDIA}
\relax{NVIDIA}, ``{Isaac {SDK}},'' 2019, available online at \url{https://developer.nvidia.com/isaac-sdk}.

\bibitem{sim2realGapEssex1995}
N.~Jakobi, P.~Husbands, and I.~Harvey, ``Noise and the reality gap: The use of simulation in evolutionary robotics,'' in \emph{European Conference on Artificial Life}.\hskip 1em plus 0.5em minus 0.4em\relax Springer, 1995, pp. 704--720.

\bibitem{artatk2022}
A.~Elmquist, A.~Young, I.~Mahajan, K.~Fahey, A.~Dashora, S.~Ashokkumar, S.~Caldararu, V.~Freire, X.~Xu, R.~Serban, and D.~Negrut, ``A software toolkit and hardware platform for investigating and comparing robot autonomy algorithms in simulation and reality,'' \emph{arXiv preprint arXiv:2206.06537}, 2022.

\bibitem{Rother2023EventTriggeredControl}
C.~Rother, Z.~Zhou, and J.~Chen, ``Development of a four-wheel steering scale vehicle for research and education on autonomous vehicle motion control,'' \emph{IEEE Robotics and Automation Letters}, vol.~8, no.~8, pp. 5015--5022, 2023.

\bibitem{pan2020imitation}
Y.~Pan, C.-A. Cheng, K.~Saigol, K.~Lee, X.~Yan, E.~A. Theodorou, and B.~Boots, ``Imitation learning for agile autonomous driving,'' \emph{The International Journal of Robotics Research}, vol.~39, no. 2-3, pp. 286--302, 2020.

\bibitem{wang2023GPlearningandMPC}
J.~Wang, M.~T. Fader, and J.~A. Marshall, ``Learning-based model predictive control for improved mobile robot path following using gaussian processes and feedback linearization,'' \emph{Journal of Field Robotics}, 2023.

\bibitem{jin2023pathfollowing}
X.~Jin, Q.~Wang, Z.~Yan, and H.~Yang, ``Nonlinear robust control of trajectory-following for autonomous ground electric vehicles with active front steering system,'' \emph{AIMS Math}, vol.~8, no.~5, pp. 11\,151--11\,179, 2023.

\bibitem{benekohal1988carsim}
R.~F. Benekohal and J.~Treiterer, ``{CARSIM}: Car-following model for simulation of traffic in normal and stop-and-go conditions,'' \emph{Transportation research record}, vol. 1194, pp. 99--111, 1988.

\bibitem{sierra2024RLpathfollowing}
J.~E. Sierra-Garcia and M.~Santos, ``Combining reinforcement learning and conventional control to improve automatic guided vehicles tracking of complex trajectories,'' \emph{Expert Systems}, vol.~41, no.~2, p. e13076, 2024.

\bibitem{zhang2021adaptiveMPCpathfollowing}
K.~Zhang, Q.~Sun, and Y.~Shi, ``Trajectory tracking control of autonomous ground vehicles using adaptive learning mpc,'' \emph{IEEE Transactions on Neural Networks and Learning Systems}, vol.~32, no.~12, pp. 5554--5564, 2021.

\bibitem{domainRandomizationAbbeel2017}
J.~Tobin, R.~Fong, A.~Ray, J.~Schneider, W.~Zaremba, and P.~Abbeel, ``Domain randomization for transferring deep neural networks from simulation to the real world,'' in \emph{2017 IEEE/RSJ international conference on intelligent robots and systems (IROS)}.\hskip 1em plus 0.5em minus 0.4em\relax IEEE, 2017, pp. 23--30.

\bibitem{voogd2023reinforcementpathfollowing}
K.~L. Voogd, J.~P. Allamaa, J.~Alonso-Mora, and T.~D. Son, ``Reinforcement learning from simulation to real world autonomous driving using digital twin,'' \emph{IFAC-PapersOnLine}, vol.~56, no.~2, pp. 1510--1515, 2023.

\bibitem{kovacs2023optimizationtubecontrol}
A.~Kovacs and I.~Vajk, ``Optimization-based model predictive tube control for autonomous ground vehicles with minimal tuning parameters,'' \emph{Unmanned Systems}, vol.~11, no.~01, pp. 93--108, 2023.

\bibitem{kadian2020sim2real}
A.~Kadian, J.~Truong, A.~Gokaslan, A.~Clegg, E.~Wijmans, S.~Lee, M.~Savva, S.~Chernova, and D.~Batra, ``{Sim2Real} predictivity: Does evaluation in simulation predict real-world performance?'' \emph{IEEE Robotics and Automation Letters}, vol.~5, no.~4, pp. 6670--6677, 2020.

\bibitem{aaronAmesSafetySim2Real2023}
P.~Akella, W.~Ubellacker, and A.~D. Ames, ``Safety-critical controller verification via sim2real gap quantification,'' in \emph{2023 IEEE International Conference on Robotics and Automation (ICRA)}.\hskip 1em plus 0.5em minus 0.4em\relax IEEE, 2023, pp. 10\,539--10\,545.

\bibitem{TR-2023-15ArtOak}
H.~Zhang, S.~Caldararu, and D.~Negrut, ``{ART-Oak},'' Simulation-Based Engineering Laboratory, University of Wisconsin-Madison, Tech. Rep., 2023, \url{https://sbel.wisc.edu/technicalreports/}.

\bibitem{chollet2015keras}
F.~Chollet \emph{et~al.}, ``Keras,'' \url{https://keras.io}, 2015.

\bibitem{paszke2017PyTorch}
A.~Paszke, S.~Gross, S.~Chintala, G.~Chanan, E.~Yang, Z.~DeVito, Z.~Lin, A.~Desmaison, L.~Antiga, and A.~Lerer, ``Automatic differentiation in {PyTorch},'' in \emph{NIPS 2017 Workshop Autodiff}, 2017, Conference Proceedings.

\bibitem{projectChronoGithub}
{Project Chrono Team}, ``{Chrono}: An open source framework for the physics-based simulation of dynamic systems,'' \url{https://github.com/projectchrono/chrono}, accessed: 2022-01-10.

\bibitem{pyChronoCondaWebSite}
\relax {Project Chrono} Development~Team, ``{PyChrono}: A {P}ython wrapper for the {C}hrono multi-physics library,'' \url{https://anaconda.org/projectchrono/pychrono}, accessed: 2023-01-14.

\bibitem{benatti2022pychrono}
S.~Benatti, A.~Young, A.~Elmquist, J.~Taves, R.~Serban, D.~Mangoni, A.~Tasora, and D.~Negrut, ``Pychrono and {G}ym-{C}hrono: A deep reinforcement learning framework leveraging multibody dynamics to control autonomous vehicles and robots,'' in \emph{Advances in Nonlinear Dynamics}.\hskip 1em plus 0.5em minus 0.4em\relax Springer, 2022, pp. 573--584.

\bibitem{chronoVehicle2019}
R.~Serban, M.~Taylor, D.~Negrut, and A.~Tasora, ``{Chrono::Vehicle} template-based ground vehicle modeling and simulation,'' \emph{International Journal of Vehicle Performance}, vol.~5, no.~1, pp. 18--39, 2019.

\bibitem{artatkResearchPlatform2022}
\BIBentryALTinterwordspacing
A.~Elmquist, A.~Young, T.~Hansen, S.~Ashokkumar, S.~Caldararu, A.~Dashora, I.~Mahajan, H.~Zhang, L.~Fang, H.~Shen, X.~Xu, R.~Serban, and D.~Negrut, ``{ART/ATK}: A research platform for assessing and mitigating the sim-to-real gap in robotics and autonomous vehicle engineering,'' 2022. [Online]. Available: \url{https://arxiv.org/pdf/2211.04886.pdf}
\BIBentrySTDinterwordspacing

\bibitem{asherSensorSimulation2021}
A.~Elmquist, R.~Serban, and D.~Negrut, ``A sensor simulation framework for training and testing robots and autonomous vehicles,'' \emph{Journal of Autonomous Vehicles and Systems}, vol.~1, no.~2, p. 021001, 2021.

\bibitem{synchrono2020}
J.~Taves, A.~Elmquist, A.~Young, R.~Serban, and D.~Negrut, ``Synchrono: A scalable, physics-based simulation platform for testing groups of autonomous vehicles and/or robots,'' in \emph{2020 IEEE/RSJ International Conference on Intelligent Robots and Systems (IROS)}.\hskip 1em plus 0.5em minus 0.4em\relax IEEE, 2020, pp. 2251--2256.

\bibitem{HarryIROSPathFollowing}
H.~Zhang, S.~Caldararu, A.~Young, A.~Ruiz, H.~Unjhawala, S.~Ashokkumar, I.~Mahajan, N.~Batagoda, L.~Bakke, and D.~Negrut, ``{Simulation Reproducibility for Different Path Following Policies},'' \url{https://github.com/uwsbel/sbel-reproducibility/tree/master/2024/IROSPathFollowing}, 2024, {Simulation-Based Engineering Laboratory, University of Wisconsin-Madison}.

\bibitem{huzaifaIEEE-AccessCalibration2024}
H.~Unjhawala, T.~Hansen, H.~Zhang, S.~Caldraru, S.~Chatterjee, L.~Bakke, J.~Wu, R.~Serban, and D.~Negrut, ``An expeditious and expressive vehicle dynamics model for applications in controls and reinforcement learning,'' \emph{IEEE Access}, pp. 1--1, 2024.

\bibitem{atk-art2022}
\relax {UW-Madison Simulation Based Engineering Laboratory}, ``{Autonomy Toolkit},'' \url{http://projects.sbel.org/autonomy-toolkit/}, 2022.

\bibitem{klanvcar2007tracking}
G.~Klan{\v{c}}ar and I.~{\v{S}}krjanc, ``Tracking-error model-based predictive control for mobile robots in real time,'' \emph{Robotics and autonomous systems}, vol.~55, no.~6, pp. 460--469, 2007.

\bibitem{osqp}
``\relax {OSQP (Operator Splitting Quadratic Program): Open-source numerical optimization solver},'' \url{https://www.cvxpy.org/}, accessed : 2023-10-21.

\bibitem{TR-2023-06}
H.~Zhang, H.~Unjhawala, S.~Caldararu, I.~Mahajan, L.~Bakke, R.~Serban, and D.~Negrut, ``Simplified {4DOF} bicycle model for robotics applications,'' Simulation-Based Engineering Laboratory, University of Wisconsin-Madison, Tech. Rep., 2023, \url{https://sbel.wisc.edu/wp-content/uploads/sites/569/2023/06/TR-2023-06.pdf}.

\end{thebibliography}


\section{APPENDIX}
\label{sec:appendix}

\begin{figure}[h!]
	\centering
	\includegraphics[width=0.48\textwidth]{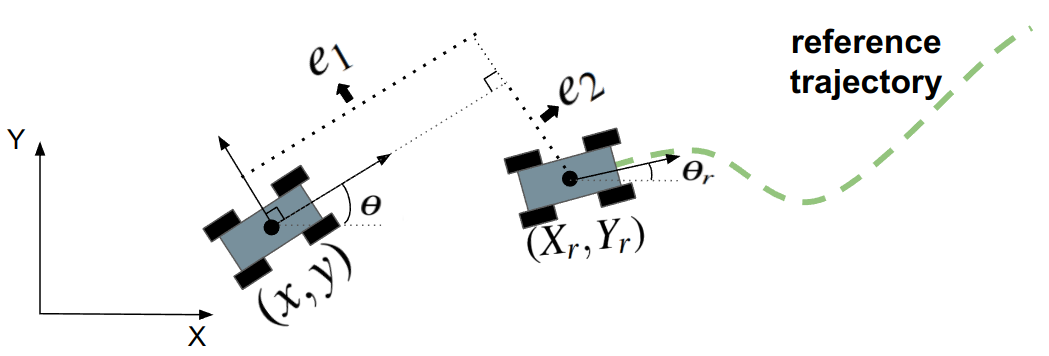}
	\caption{Error state $\mathbf e$ formulation}
	\label{fig:error_state}
\end{figure}

\noindent{\textbf{The Model Predictive Control (MPC) Solution}}. Compared to the traditional vehicle dynamics model used for MPC (often a bicycle model), the 4-DOF model employed takes in more realistic vehicle control inputs (steering and throttle command) rather than the traditional control inputs (heading, longitudinal velocity or acceleration). The state variable and control inputs are defined as $\mathbf q = [x,y,\theta,v]^T$ and $\mathbf u = [\alpha,\beta]^T$, respectively. Note that $(x,y)$ are Cartesian coordinates, $\theta$ is the heading angle with respect to the positive $x$ direction, $v$ is vehicle's longitudinal velocity, $\alpha \in [0,1]$ is the throttle command of vehicle, and $\beta \in [-1,1]$ is the steering command. To solve the reference path following problem, a predefined reference trajectory (state) $\mathbf q_r$ is given. Moreover, it is useful to define the error state $\mathbf e$ as the deviation between the current state $\mathbf q$ and the reference state $\mathbf q_r$ as shown in Eq.~(\ref{eq:error_state}) and Fig.~\ref{fig:error_state}. This error state was used in all policies in this study.
\begin{equation}
	\label{eq:error_state}
	\mathbf{e}
	=
	\begin{pmatrix}
		\cos\theta & \sin\theta & 0 & 0\\
		-\sin\theta & \cos\theta & 0 & 0\\
		0 & 0 & 1 & 0\\
		0 & 0 & 0 & 1
	\end{pmatrix}
	\begin{pmatrix}
		X_r-x\\
		Y_r-y\\
		\theta_r-\theta\\
		v_r-v
	\end{pmatrix}\;
\end{equation}

The MPC used poses a tracking-error and model based optimal control problem~\cite{klanvcar2007tracking}. Once we have the error state $\mathbf e$ and the vehicle dynamics model, we can formulate a nonlinear MPC problem based on error dynamics to perform path following. The MPC problem is formulated as solve $J_t^*(\mathbf e_t) = \min_{\mathbf u_k} \; \mathbf  e_N^T \mathbf  Q_N  \mathbf e_N +\sum_{k=0}^{N-1} \mathbf  e_k^T \mathbf Q \mathbf e_k + (\mathbf u_k-\mathbf u_r)^T \mathbf R (\mathbf u_k-\mathbf u_r)$ subject to $\mathbf e_{t+1} = \mathbf A_t \cdot \mathbf e_{t} + \mathbf B_t \cdot \mathbf u_t$.
Note that $\mathbf Q$, $\mathbf Q_N$, and $\mathbf R$ are weighting matrices in the objective function; $N$ is prediction horizon and set to be $N=10$; $\mathbf A_t$ and $\mathbf B_t$ are state transition and control matrices (as derived in the following). The optimization solver used for this work is OSQP~\cite{osqp}, and average solving time is around $6 \, \unit{ms}$.

The customized 4-DOF vehicle dynamics model is:
\begin{equation}
\label{eq:veh_dynamics}
\dot{\mathbf{q}} = \begin{pmatrix}
				\dot{x}\\
				\dot{y}\\
				\dot{\theta}\\
				\dot{v}
			\end{pmatrix}= \mathbf{f}(\mathbf{q},\mathbf{u}) = \begin{pmatrix}
                cos(\theta) \cdot v\\
                sin(\theta) \cdot v\\
                v \cdot tan(\beta \delta) \cdot l^{-1}\\
                T(\alpha,v) \cdot \gamma \cdot I_{wheel}^{-1}\cdot R_{wheel}
            \end{pmatrix}
\end{equation}
where $\mathbf{q}=[x,y,\theta,v]^T$ and $\mathbf{u}= [\alpha,\beta]^T$ are the state variable and control inputs, respectively. Note that $l$ is the distance between vehicle's front and rear axle, $\delta$ is the coefficient that maps steering command to steering angle linearly, $T(\alpha,v)$ is the empirical DC brushless motor torque function~\cite{TR-2023-06}, $\gamma$ is the gear ratio, $I_{wheel}$ is the inertia of the wheel, and $R_{wheel}$ is the radius of the wheel.

Given the vehicle dynamics as in Eq.~(\ref{eq:veh_dynamics}) and the error state formulation, the error dynamics that contains information of vehicle model and tracking-error could be derived by plugging  Eq.~(\ref{eq:veh_dynamics}) in Eq.~(\ref{eq:error_state}) as
\begin{equation}
	\dot{\mathbf e} = 
	\begin{pmatrix}
		\frac{v \cdot \tan(\delta \beta) \cdot e_2}{l}+v_r \cdot \cos{e_3} -v\\
		-\frac{v \cdot \tan(\delta \beta) \cdot e_1}{l} + v_r \cdot \sin{e_3}\\
		\frac{v_r\cdot \tan(\delta_r \beta) - v \cdot \tan(\delta \beta)}{l}\\
		\frac{\tau_0 R_{wheel}  \gamma}{I_w}(\alpha_r-\alpha)-e_4 \,   \frac{c_1 \omega_0 + \tau_0 }{I_w \omega_0}
	\end{pmatrix}
	= \mathbf g(\mathbf e,\mathbf u)
\label{eq:err_dyn_c}
\end{equation}
Note the unseen variable $c_1, \tau_0$ are from $T(\alpha, v)$ in Eq.~(\ref{eq:veh_dynamics}). To use the error dynamics model in MPC problem, we need to linearize the Eq.~(\ref{eq:err_dyn_c}) around optimal error state $\mathbf e^*$ and control input $\mathbf u^*$ as below (notice that $\mathbf e^*$ and $\mathbf u^*$ does not have to be zero vectors because of the non-zero look ahead distance and non-zero throttle to maintain constant velocity):
\begin{equation}
	\label{equ:error_d_continous}
	\begin{split}
		&\dot{\mathbf e} = \frac{\partial \mathbf g}{\partial \mathbf e}_{\;\; (\mathbf{e^*\; u^*})} \cdot \mathbf e + \frac{\partial \mathbf g}{\partial \mathbf u}_{\;\; (\mathbf{e^*\; u^*})} \cdot \mathbf u =\\
		&\begin{pmatrix}
			0 & \frac{v \cdot \tan(\delta \beta)}{l} & -v_r \cdot \sin{e_3} & 0 \\
			-\frac{v \cdot \tan(\delta \beta)}{l} & 0 & v_r \cdot \cos{e_3} & 0 \\
			0 & 0 & 0 & 0 \\
			0 & 0 & 0 & -\frac{c_1 \omega_0 + \tau_0}{I_{wheel} \omega_0} 
		\end{pmatrix}_{ (\mathbf{e^*\; u^*})}\\
		& \mathbf e 
		+
		\begin{pmatrix}
			0 & \frac{v \cdot e_2 \delta}{l \cdot \cos^2(\beta \delta)}\\
			0 & -\frac{v \cdot e_1 \delta}{l \cdot \cos^2(\beta \delta)} \\
			0 & -\frac{v \delta}{l \cdot \cos^2(\beta \delta)}\\
			-\frac{\tau_0 R_{wheel}  \gamma}{I_{wheel}} & 0
		\end{pmatrix}_{(\mathbf{e^*\; u^*})}
		 \mathbf  u =\mathbf A_c \; \mathbf e + \mathbf B_c \; \mathbf u
	\end{split}
\end{equation}
Then we discretize Eq.~(\ref{equ:error_d_continous}) using time step $\Delta t = 0.1 \unit{s}$ in the following form: 
\begin{equation}
	\label{equ:error_state_discrete}
	\mathbf e_{t+1} = \mathbf A_t \cdot \mathbf e_{t} + \mathbf B_t \cdot \mathbf u_t 
\end{equation}    
\begin{equation*}
	\mathbf A_t =\mathbf A_c \; \Delta t + \mathbb{I}_{4 \times 4}\;\;\;\;\;\;
	\mathbf B_t = \mathbf B_c \; \Delta t
\end{equation*}

\medskip

\noindent{\textbf{Proportional-Integral-Derivative (PID) Controller}}. The PID controller is a simple feedback control policy that widely used in controls engineering. The PID controller is designed to minimize the error state $\mathbf e$ by adjusting the control inputs $\mathbf u$ based on the error state. The PID parameter tuning process could be replaced by a simple least square optimization once the training data is available. The training data used is the same as the training data for NN-MPC (see Sec.~\ref{sec:method}). The coefficient is obtained by solving the following least square optimization problem: $\min_{\mathbf c} \; || \mathbf{E c - U} ||^2 $, where $\mathbf{E}$ and $\mathbf{U}$ are the error state and control input training data, and $\mathbf{c}$ is the coefficient of PID controller. The average solving time for PID controller was around $0.05 \, \unit{ms}$.

\end{document}